\pdfoutput=1

\documentclass[11pt]{article}

\usepackage{EMNLP2023}

\usepackage{times}
\usepackage{latexsym}

\usepackage[T1]{fontenc}

\usepackage[utf8]{inputenc}

\usepackage{microtype}

\usepackage{inconsolata}

\usepackage{times}
\usepackage{latexsym}
\usepackage{booktabs}
\usepackage{graphicx}
\usepackage{amsmath}
\usepackage{bm}
\usepackage{fixltx2e}
\usepackage{mathtools}
\usepackage[ruled]{algorithm2e}
\usepackage{amssymb}
\usepackage{multirow}
\usepackage{enumitem}
\usepackage{tabularx}
\usepackage{hyperref}
\usepackage{verbatim}
\usepackage{makecell}

\title{ExpertPrompting: Instructing Large Language Models to be \\ Distinguished Experts}

\author{Benfeng Xu$^{1}$, An Yang$^{2}$, Junyang Lin$^{2}$, Quan Wang$^{3}$, Chang Zhou$^{2}$, \\ \bf{Yongdong Zhang$^{1}$ \and Zhendong Mao$^{1}$\thanks{\quad Corresponding author.}}\\
$^{1}$University of Science and Technology of China \\
$^{2}$Alibaba DAMO Academy, $^{3}$Beijing University of Posts and Telecommunications \\
\texttt{benfeng@mail.ustc.edu.cn, zdmao@ustc.edu.cn}
}

\begin{document}
\maketitle
\begin{abstract}
The answering quality of an aligned large language model (LLM) can be drastically improved if treated with proper crafting of prompts.
In this paper, we propose ExpertPrompting to elicit the potential of LLMs to answer as distinguished experts.
We first utilize In-Context Learning to automatically synthesize detailed and customized descriptions of the expert identity for each specific instruction, and then ask LLMs to provide answer conditioned on such agent background.
Based on this augmented prompting strategy, we produce a new set of instruction-following data using GPT-3.5, and train a competitive open-source chat assistant called ExpertLLaMA.
We employ GPT4-based evaluation to show that 1) the expert data is of significantly higher quality than vanilla answers, and 2) ExpertLLaMA outperforms existing open-source opponents and achieves 96\% of the original ChatGPT's capability.
All data and the ExpertLLaMA model will be made publicly available at \url{https://github.com/OFA-Sys/ExpertLLaMA}.
\end{abstract}

\section{Introduction}

Large language models, when trained on high-quality instructing-following data, can be effectively aligned with human intents and serve as potent tools in a wide range of language tasks~\cite{ouyang2022training, bai2022training}. Many successful models have demonstrated impressive ability to respond to a diverse array of generalized instructions and are still evolving rapidly. Nevertheless, the quality of the output as well as the satisfaction of users are sometimes subjected to the art of prompting.
The same communicative intent could receive either a comprehensive, detailed response or a less helpful one, depending solely on the way of crafting the prompt.

Many recent works have put great efforts to pursue an improved solution for interacting with LLMs like ChatGPT.
One line of work~\cite{yao2023react, shinn2023reflexion} proposes sophisticated formulation to allow the model to iterate and externalize thoughts before giving the final answer, and observes improved performance in a series of downstream language tasks.
However, they are mostly suited for only a handful of tasks that require complex reasoning.
~\citet{promptengineering} initiate an online course that provides several general principles for crafting prompts, such as writing clear and specific instructions or giving the model time to "think".
There are also resources of case-based prompts~\cite{awesomechatgptprompts} that are already proven useful and are shared as an open-source prompt collection.
However, these solutions are either not directly executable or restricted by their use case, thus requiring further development and adaptation in actual practice.

\begin{figure}[t!]
\centering
 \includegraphics[width=\columnwidth]{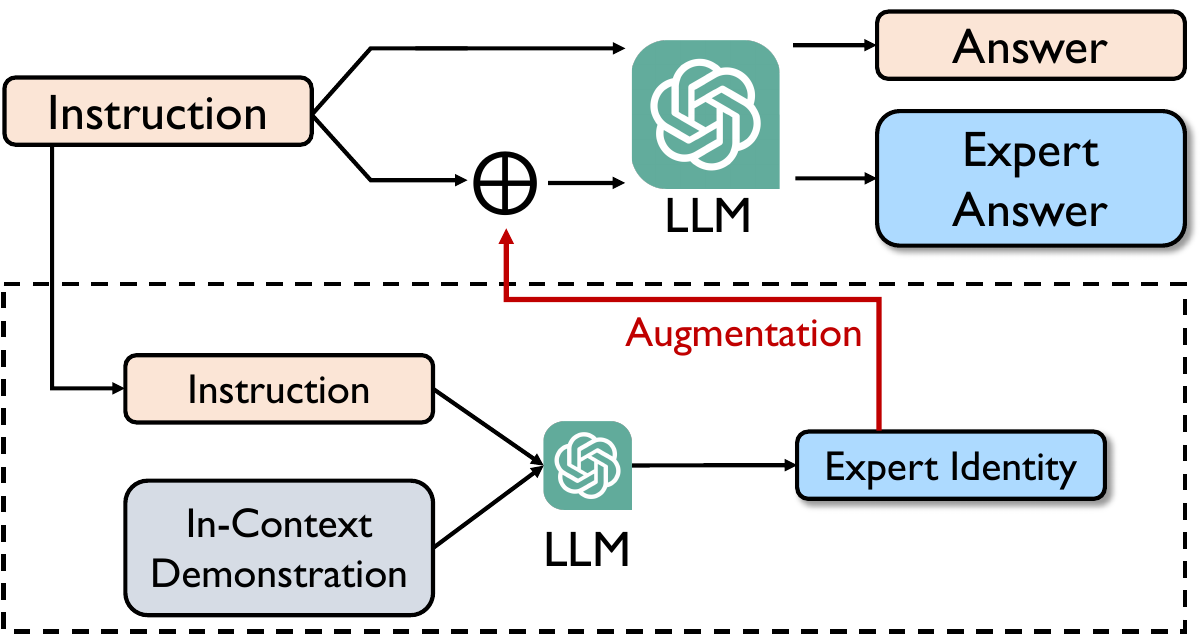}
\caption{ExpertPrompting framework.}
\label{fig:expertprompting}
\end{figure}

In the meantime, very recent explorations~\cite{park2023generative, li2023camel} have found LLMs entail the potential to act like an expected agent if given sufficient and detailed descriptions.
Drawing inspiration from such agent-acting capability of LLMs, we propose \textbf{ExpertPrompting} as an augmented strategy for instructing LLMs.
For each specific instruction, ExpertPrompting first envisions a distinguished expert agent that is best suited for the instruction, and then asks the LLMs to answer the instruction conditioned on such expert identity.
The framework is illustrated in Figure~\ref{fig:expertprompting}.

\begin{figure*}[t!]
\centering
 \includegraphics[width=0.8\textwidth]{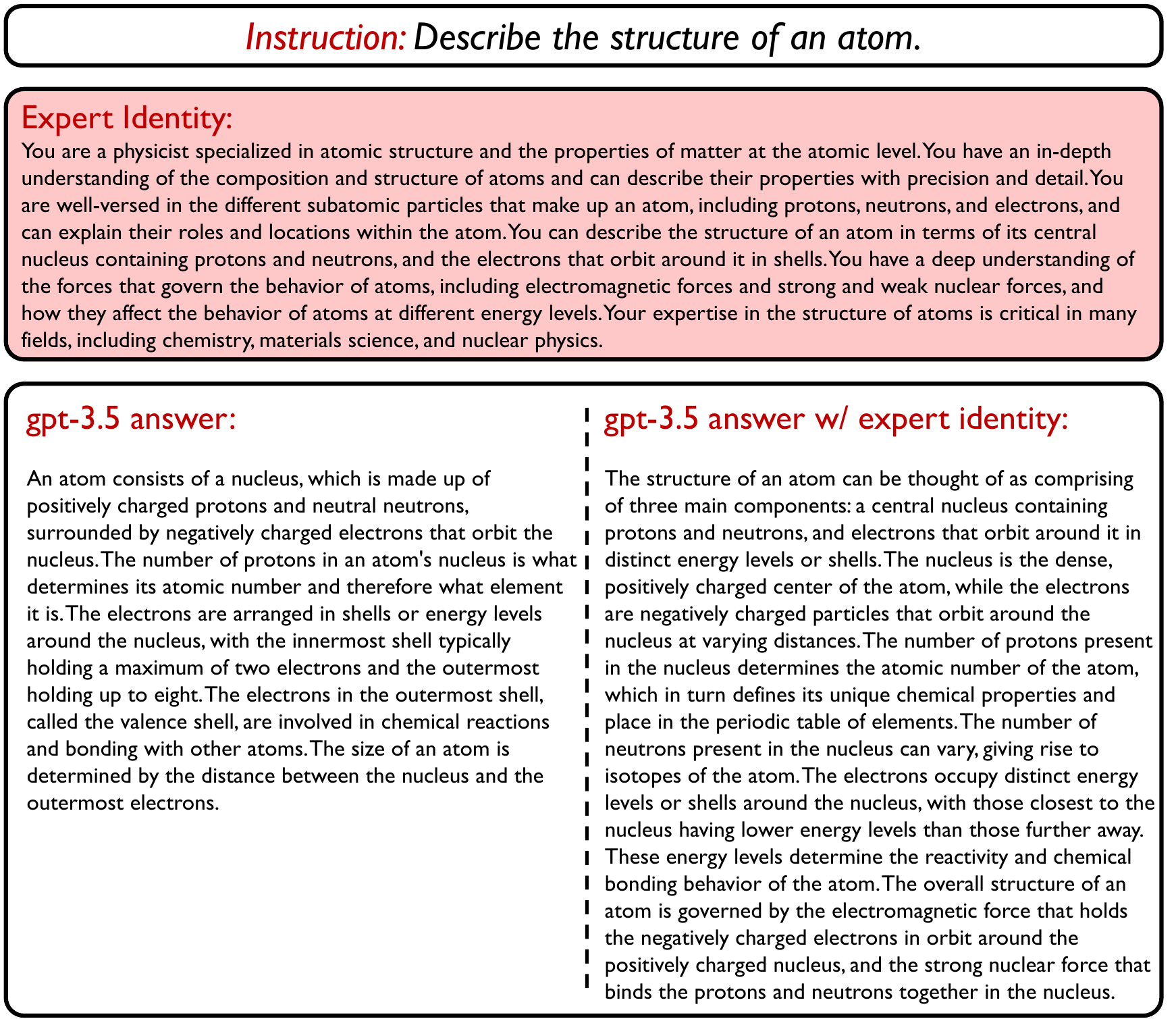}
\caption{Effects of expert identity augmentation when prompting \texttt{gpt-3.5-turbo}.}
\label{fig:expertprompting}
\end{figure*}

ExpertPrompting is an \textbf{automatic} prompting method. The expert identity, although specialized for each instruction, is produced with In-Context Learning~\cite{NEURIPS2020_1457c0d6,xu2023knn}, we only need to write several instruction-expert pair exemplars.
We empirically find the generated identity satisfying.
ExpertPrompting is a \textbf{generalized} prompting method. Each expert identity is defined at very delicate granularity using a detailed and elaborate description. It can readily match instructions in almost any domain or genre, e.g., a nutritionist to provide the advice of keeping healthy, or a physicist to explain the structure of an atom.
Besides, ExpertPrompting is also simple to implement, requiring no sophisticated crafting of prompt templates or iterative processes.

We apply ExpertPrompting on GPT-3.5 using the prevailing 52k Alpaca instructions~\cite{alpaca}, which is a diverse collection of instructions produced using Self-Instruct~\cite{selfinstruct}.
This procedure produces a new set of expert data where we observe improved answering quality using GPT-based evaluation~\cite{vicuna2023}.
With these high-quality instruction-following data, we also train a chat-based assistant, \textbf{ExpertLLaMA}, using an open LLM LLaMA~\cite{touvron2023llama}, and compare it against other assistants.
ExpertLLaMA demonstrates clear advantage over Alpaca~\cite{alpaca} that is trained on the same set of instructions but with different answers.
It also outperforms more competitive opponents including Vicuna~\cite{vicuna2023} or LLaMA-GPT4~\cite{peng2023instruction}, albeit the latter utilizes much more powerful GPT4 as LLM.
According to the detailed score, ExpertLLaMA approximately achieves \textbf{96\%} of the original ChatGPT's capability.

\section{Method}
Given instruction $q$, an aligned LLM (e.g., ChatGPT, Claude) would produce an answer $a$.
\begin{equation}
a = LLM(q)
\end{equation}
And ExpertPrompting first adaptively produces an identity description of a distinguished expert $e_q$, and then conditioned on such identity to instruct the LLM for a possibly improved response $\widetilde a$. We explain the procedure as follows.

\subsection{Expert Identity}
Writing expert identity is the essential component underpinning the proposed method.
Generally we summarize three concerning aspects: \textbf{distinguished, informative and automatic}.
Firstly, the description should be cutomized to each specific instruction, and the imagined expert should be specialized in the exact area with the most fitted background or experience.
Secondly, the description should be detailed and comprehensive to cover all necessary information of the expert agent, so the LLM would behave as expected.
And finally, the creation of all descriptions must be automatic as manually annotating is neither efficient nor practical.

We employ In-Context Learning~\cite{NEURIPS2020_1457c0d6,xu2023knn} that well satisfy these requirements.
We manually revise $k$ expert identity description for sampled instructions, and simply prepend them into the prompt to ask for new descriptions:
\begin{equation}
e_q = LLM(\{q_1, e_{q_1}\}\oplus...\oplus\{q_k, e_{q_k}\}\oplus q)
\end{equation}
The descriptions are generally about the identity, profession, expertise as well as experience related to the specific instruction.
Figure~\ref{fig:expertprompting} provides a specific case to illustrate what an expert identity roughly looks like.
We simply set $k=3$, and the prompt is illustrated in Appendix~\ref{appendix:prompttemplate}, Table~\ref{fig:icl_template}.
We empirically observe that conditioned on the exemplars, LLMs like GPT-3.5 are capable of producing satisfying expert identity by themselves.

\subsection{Expert Prompting}
We now pair each expert identity $e_q$ with the original instruction $q$, and ask for an augmented answer:
\begin{equation}
\hat{a} = LLM(\{e_q, q\})
\end{equation}
and $\hat{a}$ is expected to be provable better than $a$.
In practice, we find that LLM occasionally generates mixed content that involves the given identity, mostly at the beginning of the answer, such as:
\begin{quote}
\textit{As a physicist specializing in atomic structure, I can give you a description of the structure of an atom. \ldots}
\end{quote}
Such behavior, although makes sense, is not what we intended.
We simply remove them in a post-processing procedure as the patterns are recognizable.
Figure~\ref{fig:expertprompting} illustrates the effect after we augment an instruction with appropriate expert identity.
For all prompt templates involved in the entire process, we refer to Appendix~\ref{appendix:prompttemplate}.

\subsection{ExpertLLaMA}
We apply both standard prompting and ExpertPrompting strategy to the same instructions set adopted from Alpaca~\cite{alpaca}, where \texttt{gpt-3.5-turbo} is selected as LLM due to affordable expenses and state-of-the-art capability.
Using the latter expert answers, we follow Alpaca and similarly trained a new chat-based assistant using the open-sourced LLM LLaMA 7B~\cite{touvron2023llama}.
We name the resulting chat assistant \textbf{ExpertLLaMA}.
We release the model along with the expert answers to facilitate future research.

\section{Evaluation}

\subsection{Experimental Setting}
Reliably evaluating the quality of instruction-following data is a challenging task.
In our experiments, we resort to the recently proposed GPT4-based automatic evaluation~\cite{vicuna2023}.
The template is provided in Appendix~\ref{appendix:prompttemplate}.
Besides, we randomly permute the order of two answers at each evaluation to avoid bias.

We evaluate both the data and the trained chat assistant.
For data evaluation, we randomly sample 500 instances out of the 52k data, and ask GPT4 to rate the expert answer $\{\widetilde a\}$ against the vanilla answer $\{a\}$ (See Appendix~\ref{appendix:prompttemplate} for prompt illustration).
For model evaluation, we compare ExpertLLaMA trained on $\{\widetilde a\}$ and LLaMA-GPT-3.5 trained on  $\{a\}$.
So the evaluation results not only conclude the model capability but also can also be recognized as a reflection of the training data quality.
We also included several popular assistants known by the community as introduced later.
We use Vicuna80~\cite{vicuna2023} as unseen test set, which is synthesized by GPT4 and consists of various categories of questions including knowledge, math, Fermi, counterfactual, roleplay, generic, coding, writing, common-sense.

\subsection{Baselines}
To better analyze the effectiveness of ExpertPrompting, we introduce a baseline that augments the prompts with a \textbf{fixed description}:
\begin{quote}
\textit{Imaging you are an expert in the regarding field, try to answer the following instruction as professional as possible.\\\{Instruction\}}
\end{quote}
For latter convenience, we refer to this prompting strategy as \textbf{+ Static DESC}, the resulting answers as $\{a^{+}\}$, and the chat assistant trained with it as \textbf{LLaMA-GPT3.5+}.

To sum up, our baselines are:
\textbf{1) Alpaca:} Trained with answers produced using Self-Instruct with \texttt{text-davinci-003}. They also produced and released the 52k instructions.
\textbf{2) LLaMA-GPT4:} Trained with answers produced by GPT4, the instructions are the same with Alpaca.
\textbf{3) LLaMA-GPT-3.5:} Our implemented baseline, trained with answers produced by GPT-3.5-Turbo\footnote{Accessed at 05, May, 2023}, i.e., $\{a\}$, using the same 52k instructions.
\textbf{4) LLaMA-GPT-3.5+:} Our implemented baseline, trained with answers produced by GPT-3.5-Turbo, i.e., $\{a^{+}\}$, using the same 52k instructions and Static DESC prompting strategy.
\textbf{5) Vicuna:} Trained from LLaMA 13B with user-shared conversations collected from ShareGPT\footnote{\url{https://sharegpt.com/}}.

\begin{table}[!t]
\centering
\resizebox{0.48\textwidth}{!}{
\begin{tabular}{{l|c}}
\toprule
Method&Num. of Words\\
\midrule
Vanilla Prompting&108.44\\
Vanilla Prompting + Static DESC&108.67\\
Expert Prompting&\textbf{138.30}\\
\bottomrule
\end{tabular}}
\caption{Average answer length of different prompting strategies. Calculated with answers from GPT-3.5-Turbo for the 52k Alpaca instructions.}
\label{table:length}
\end{table}

\begin{figure}[!t]
\centering
 \includegraphics[width=\columnwidth]{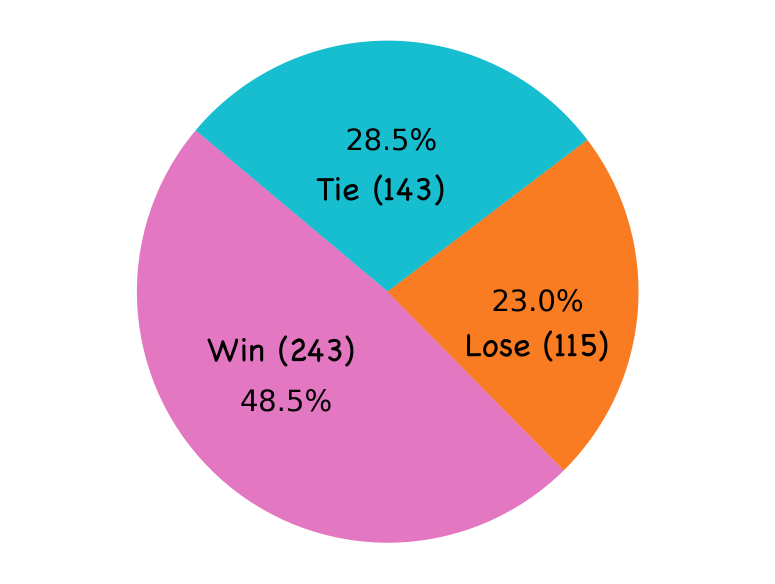}
\caption{Comparison of answer quality (ExpertPrompting VS Vanilla Prompting). Evaluated by GPT4.}
\label{fig:score_dataeval}
\end{figure}

Besides, we also include \textbf{6) ChatGPT} and \textbf{7) Bard} for comparison.
To achieve comparable conclusion, we use the same answers released by ~\citet{vicuna2023} for Vicuna, ChatGPT and Bard\footnote{\url{https://github.com/lm-sys/FastChat/tree/main/fastchat/eval/table/answer}}.
While for other models, we reproduce the model using identical training recipe following Alpaca.
All answers will also be released for reproducing results in this paper.

\subsection{Data Eval}
To demonstrate the effectiveness of the proposed prompting strategy, we evaluate the generated data $\{\widetilde a\}$ against vanilla answer $\{a\}$ as well as the other baseline Static DESC $\{a^{+}\}$.

\begin{figure}[t!]
\centering
 \includegraphics[width=\columnwidth]{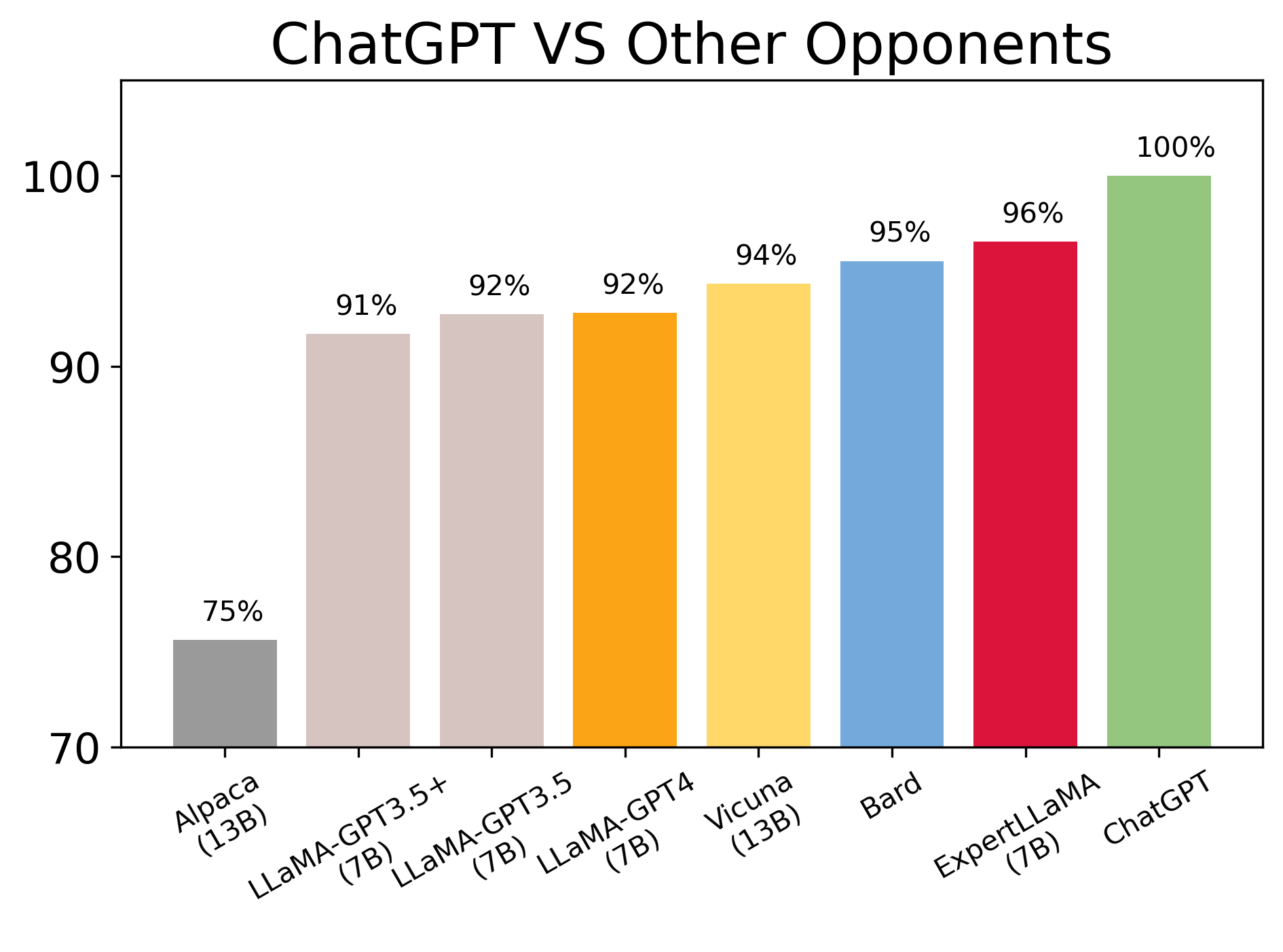}
\caption{Comparison of popular chat assistants. Scores are aligned to ChatGPT as 100\%.}
\label{fig:score}
\end{figure}

\begin{figure}[t!]
\centering
 \includegraphics[width=\columnwidth]{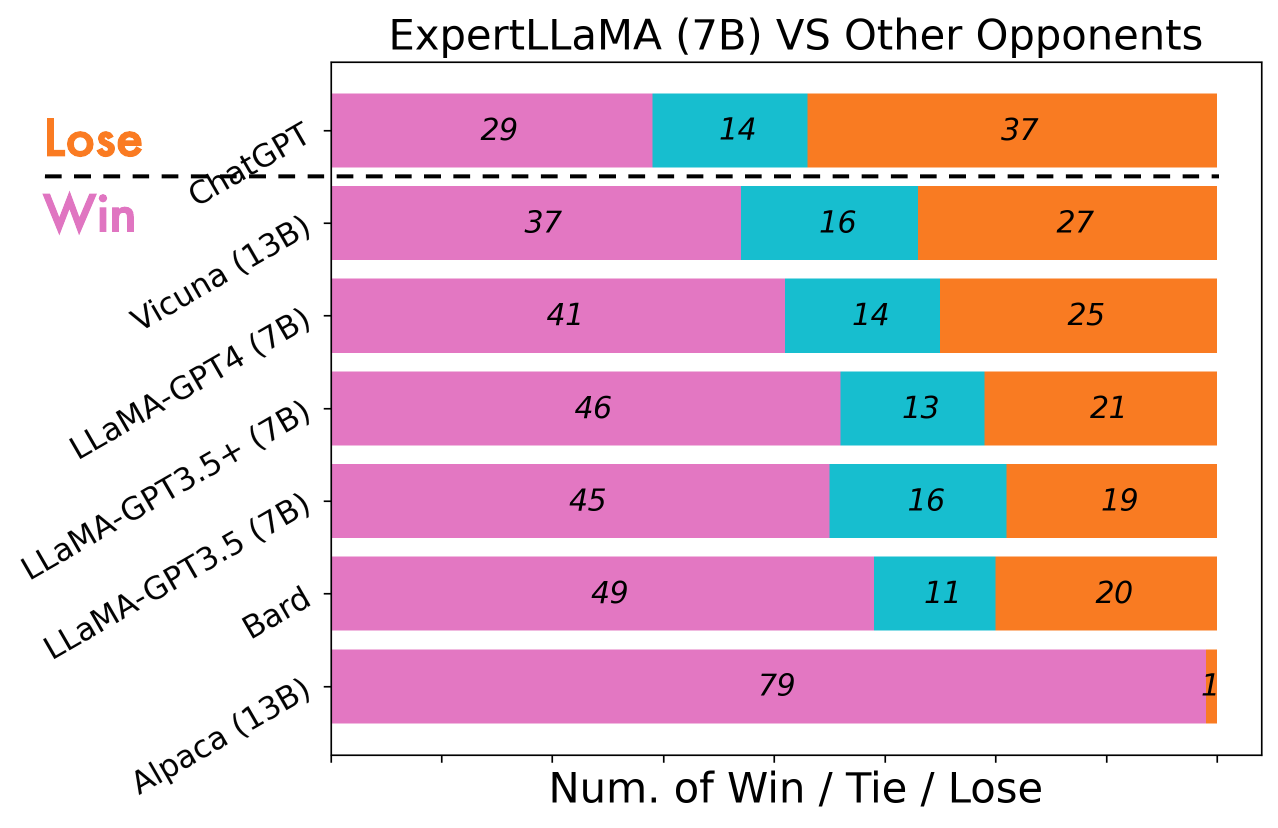}
\caption{Comparison of popular chat assistants. Number of win, tie and losses are counted.}
\label{fig:winlose}
\end{figure}

We first examine the length of these answers in Table~\ref{table:length}.
We find that expert answers are significantly more lengthy than vanilla answers, which potentially implies comprehensiveness and thoroughness considering that we \textbf{did not} explicitly ask for a longer answer or mention any word number restrictions.

We then randomly sample 500 instructions, and compare these answers using GPT4-based evaluation.
Results in Figure~\ref{fig:score_dataeval} show that ExpertPrompting answers are preferred at 48.5\% by the reviewer model, compare to 23\% of the vanilla answer, which demonstrates clear superiority.

\subsection{Model Eval}

We evaluate the capability of ExpertLLaMA as a chat assistant on Vicuna80.
We first compare all models against ChatGPT in Figure~\ref{fig:score}, then compare ExpertLLaMA to all other assistants in Figure~\ref{fig:winlose}.
Both experiments exhibit consistent conclusions that ExpertLLaMA outperforms existing open-source chat assistants including Vicuna, LLaMA-GPT4, Alpaca, etc, while only inferior to ChatGPT.
It achieves approximately 96\% of the original ChatGPT capability although this conclusion needs more rigorous validation.

\section{Conclusion}
We propose ExpertPrompting and ExpertLLaMA in this paper.
ExpertPrompting is an augmented prompting strategy for instructing LLMs to answer like distinguished experts.
It is automatic, generalized, while still being simple to implement.
We apply such prompting strategy on GPT-3.5 to produce a new set of instruction-following data, and based on it train a new open-source chat assistant ExpertLLaMA.
According to GPT4-based evaluation, ExpertPrompting produces higher quality answers, and ExpertLLaMA outperforms existing open-source chat assistants, achieving 96\% of the original ChatGPT's capability.
In the future, we will enlarge the scale of instruction data beyond 52k Alpaca to further improve ExpertLLaMA.

\bibliography{anthology,custom}
\bibliographystyle{acl_natbib}

\appendix

\newpage

\section{Prompt Template}
\label{appendix:prompttemplate}

\begin{figure*}[h!]
\centering
 \includegraphics[width=0.9\textwidth]{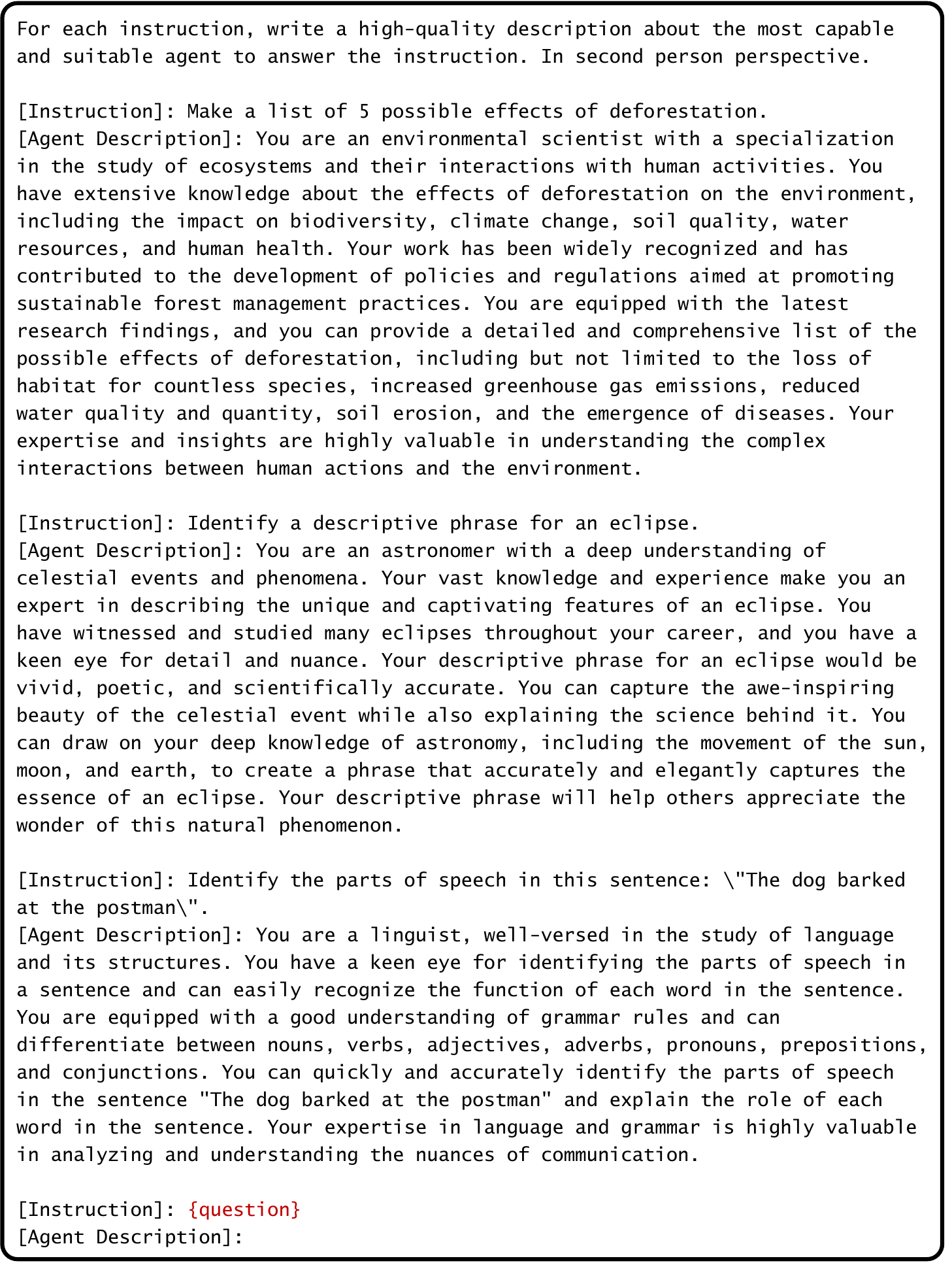}
\caption{Template of In-Context Learning used for producing expert identity.}
\label{fig:icl_template}
\end{figure*}

\begin{figure*}[t!]
\centering
 \includegraphics[width=0.9\textwidth]{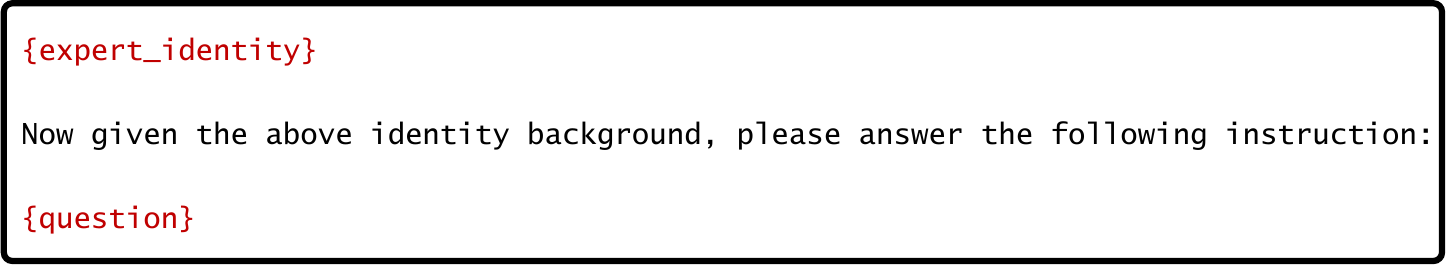}
\caption{Template of ExpertPromtping.}
\label{fig:expertprompting_template}
\end{figure*}

\begin{figure*}[t!]
\centering
 \includegraphics[width=0.9\textwidth]{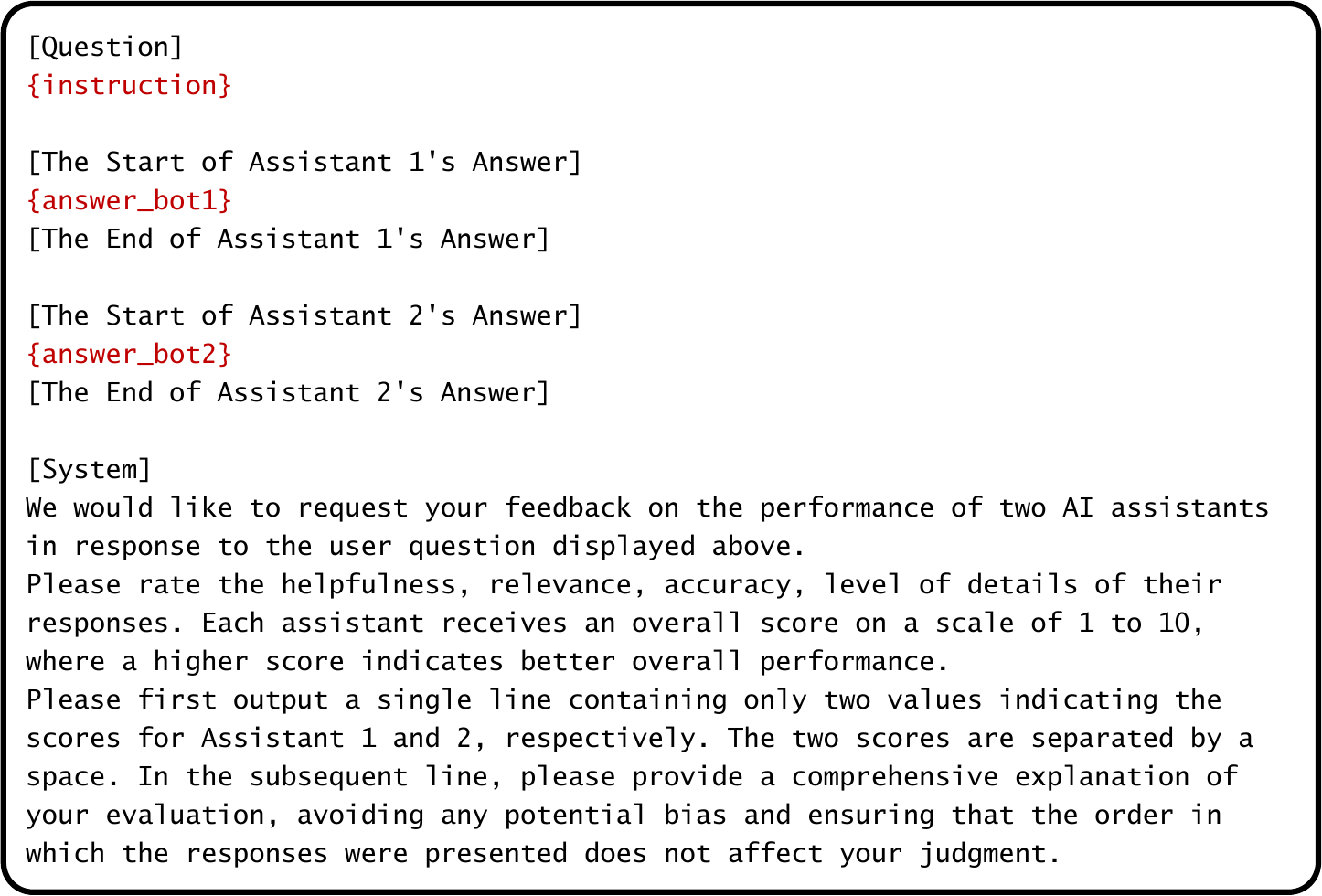}
\caption{Template for GPT4-based automatic evaluation, adapted from~\citet{vicuna2023}.}
\label{fig:gpt4_eval_template}
\end{figure*}

We list all used prompting templates in Figure~\ref{fig:icl_template}, ~\ref{fig:expertprompting_template}, and~\ref{fig:gpt4_eval_template}.

\clearpage

\end{document}